\documentclass{sig-alternate-05-2015}

\usepackage{color}
\usepackage[dvipsnames]{xcolor}

\newcommand{\support}{\mathrm{support}}
\newcommand{\confidence}{\mathrm{conf}}

\newcommand{\causes}{\rightarrow}
\newcommand{\att}{\mathrm{ATT}}
\newcommand{\E}{\mathbb{E}}
\newcommand{\experiment}[1]{\vspace{2mm}\noindent{\textit{#1}}}
\newcommand{\prob}{\mathrm{P}}

\newcommand{\myparag}[1]{\vspace{1mm}\noindent\textbf{#1}}
\newcommand{\mysubparag}[1]{\vspace{1mm}\noindent\textit{#1}}

\newtheorem{lemma}{Lemma}
\newtheorem{proposition}{Proposition}
\newtheorem{defn}{Definition}

\begin{document}
\onecolumn

% Copyright
\setcopyright{acmcopyright}
%\setcopyright{acmlicensed}
%\setcopyright{rightsretained}
%\setcopyright{usgov}
%\setcopyright{usgovmixed}
%\setcopyright{cagov}
%\setcopyright{cagovmixed}

%\CopyrightYear{2007} % Allows default copyright year (20XX) to be over-ridden - IF NEED BE.
%\crdata{0-12345-67-8/90/01}  % Allows default copyright data (0-89791-88-6/97/05) to be over-ridden - IF NEED BE.
% --- End of Author Metadata ---

\title{Causal Inference in Observational Data }

\numberofauthors{8} %  in this sample file, there are a *total*
\author{
% 1st. author
\alignauthor
Pranjul Yadav \\
       \affaddr{Dept. of Computer Science}\\
       \affaddr{University of Minnesota}\\
       \affaddr{Minneapolis, MN}\\
       \email{yadav023@umn.edu}
% 2nd. author
\alignauthor
Michael Steinbach \\
       \affaddr{Dept. of Computer Science}\\
       \affaddr{University of Minnesota}\\
       \affaddr{Minneapolis, MN}\\
       \email{stei0062@umn.edu}
% 6th. author
\alignauthor
Vipin Kumar \\
        \affaddr{Dept. of Computer Science}\\
       \affaddr{University of Minnesota}\\
       \affaddr{Minneapolis, USA}\\
       \email{kumar001@umn.edu}
\and
% 4th. author
\alignauthor
Bonnie Westra \\
       \affaddr{School of Nursing}\\
       \affaddr{University of Minnesota}\\
       \affaddr{Minneapolis, MN}\\
       \email{westr006@umn.edu}
% 6th. author
\alignauthor
Alexander Hoff \\
        \affaddr{Dept. of Computer Science}\\
       \affaddr{University of Minnesota}\\
       \affaddr{Minneapolis, USA}\\
       \email{hoffx231@umn.edu}
% 4th. author
\alignauthor
Connie Delaney \\
       \affaddr{School of Nursing}\\
       \affaddr{University of Minnesota}\\
       \affaddr{Minneapolis, USA}\\
       \email{delaney@umn.edu}
\and
\alignauthor
Lisiane Prunelli \\
        \affaddr{School of Nursing}\\
       \affaddr{University of Minnesota}\\
       \affaddr{Minneapolis, USA}\\
       \email{pruin001@umn.edu}
% 4th. author
\alignauthor
Gyorgy Simon \\
       \affaddr{Dept. of Health Sciences  Research}\\
       \affaddr{Mayo Clinic, Rochester, MN}\\
       \email{simon.gyorgy@mayo.edu}
}

\maketitle
\begin{abstract}
Our aging population increasingly suffers from multiple chronic diseases simultaneously, necessitating the comprehensive treatment of these conditions.  Finding the optimal set of drugs and interventions for a combinatorial set of diseases is a combinatorial pattern exploration problem. Association rule mining is a popular tool for such problems, but the requirement of health care for finding causal, rather than associative, patterns renders association rule mining unsuitable.  One of the purpose of this study was to apply SSC guideline recommendations to EHR data for patients with severe sepsis or septic shock and determine guideline compliance as well as its impact on inpatient mortality and sepsis complications. Propensity Score Matching in conjuction with Bootstrap Simulation were used to match patients with and without exposure to the SCC recommendations. Findings showed that EHR data could be used to estimate compliance with SCC recommendations as well as the effect of compliance on outcomes. Further,  we propose a novel framework based on the Rubin-Neyman causal model for extracting causal rules from observational data, correcting for a number of common biases. Specifically, given a set of interventions and a set of items that define subpopulations (e.g., diseases), we wish to find all subpopulations in which effective intervention combinations exist and in each such subpopulation, we wish to find all intervention combinations such that dropping any intervention from this combination will reduce the efficacy of the treatment. A key aspect of our framework is the concept of closed intervention sets which extend the concept of quantifying the effect of a single intervention to a set of concurrent interventions.  Closed intervention sets also allow for a pruning strategy that is strictly more efficient than the traditional pruning strategy used by the Apriori algorithm. To implement our ideas, we introduce and compare five methods of estimating causal effect from observational data and rigorously evaluate them on synthetic data to mathematically prove (when possible) why they work. We also evaluated our causal rule mining framework on the Electronic Health Records (EHR) data of a large cohort of patients from Mayo Clinic and showed that the patterns we extracted are sufficiently rich to explain the controversial findings in the medical literature regarding the effect of a class of cholesterol drugs on Type-II Diabetes Mellitus (T2DM).\\
\end{abstract}


\section{Related Work}

% Causal work
Causation has received substantial research interest in many areas. In computer science,
Pearl \cite{didelez2001judea} and Rosenbaum\cite{rosenbaum2002observational} laid the foundation for causal inference, upon which several fields, cognitive science, econometrics, epidemiology, philosophy and statistics have built their respective methodologies \cite{freedman1997association,bielby1977structural, robins2000marginal}. 

% Causal model
At the center of causation is a causal model. Arguably, one of the earliest and popular models is the
Rubin-Neyman causal model \cite{sekhon2008neyman}. Under this model $X$ causes $Y$, if $X$ 
occusr before $Y$; and without $X$, $Y$ would be different. 
Beside the Rubin-Neyman model, there are several other causal models, including the Granger causality
\cite{granger1988some} for time series, Bayes Networks \cite{jensen1996introduction}, 
Structural Equation Modeling \cite{bielby1977structural}, causal graphical models \cite{elwert2013graphical}, and more generally,
probabilistic graphical models \cite{murphy2002dynamic}. 
In our work, we use the potential outcome framework from the Rubin-Neyman model and we use
causal graphical models to identify and correct for biases.

Causal graphical models are tools to visualize causal relationships among variables.
Nodes of the causal graph are variables and edges are causal relationships. 
Most methods assume that the causal graph structure is a priori given, however,
methods have been proposed for discovering the structure of the causal graph \cite{heckerman1995bayesian, heckerman1997bayesian}. In our work, the structure is partially given: we know the relationships
among groups of variables, however we have to assign each variable to the correct group based on data.

Knowing the correct graph structure is important, because substructures in the graph are suggestive
of sources of bias. To correct for biases, we are looking for specific substructures.  For example,
causal chains can be sources of overcorrection bias and "V"-shaped structures can
be indicative of confounding or endogenous selection bias \cite{robins2000marginal}. 
Many other interesting substructures have been studied  \cite{cooper1997simple,silverstein2000scalable,mani2012theoretical}.
In our work, we consider three fundamental such structures: direct causal effect, indirect causal effect
and confounding.  Of these, confounding is the most severe and received the most research interest.

Numerous methods exist to handle confounding, which includes propensity score matching (PSM)
\cite{austin2011introduction}, structural marginal models \cite{robins2000marginal} and g-estimation \cite{bielby1977structural}.
The latter two extend PSM for various situations, for example, for time-varying interventions \cite{robins2000marginal}.

Propensity score matching is used to estimate the effect of an intervention on an outcome.
The propensity score is the propensity (probability) of a patient receiving the intervention given
his baseline characteristics and the propensity score is used to create a new population that
is free of confounding.  Many PSM techniques exist and they typically differ in how they
use the propensity score to create this new population \cite{lunceford2004stratification,austin2015estimating,
rosenbaum1983central,austin2015moving}.

Applications of causal modeling is not exclusive to social and life sciences. In data mining,
Lambert et al. \cite{lambert2007more} investigated the causal effect of new features on click through rates
and Chan et al. \cite{chan2010evaluating} used doubly robust estimation techniquest to determine the efficacy of display advertisements.

Even extending association rules mining to causal rule mining has been attempted before \cite{li2013mining,
holland1988differential,li2015observational}. 
Li et al. \cite{li2013mining} used odds ratio to identify causal patterns and 
later extended their technique \cite{li2015observational} to handle large data set. Their technique, however,
is not rooted in a causal model and hence offers no protection against computing systematically biased 
estimates. In their proposed causal decision trees \cite{li2015causal}, they used the potential outcome
framework, but still have not addressed correction for various biases, including confounding.

\section{Simple Causal Rule Mining in  Irregular Time-Series Data}

\subsection{Introduction}
According to the Center for Disease Control and Prevention, the incidence of sepsis or septicemia has doubled from 2000 through 2008, and hospitalizations have increased by 70\% for these diagnoses1. In addition, severe sepsis and shock have higher mortality rates than other sepsis diagnoses, accounting for an estimated mortality between 18\% and 40\%. During the first 30 days of hospitalization, mortality can range from 10\% to 50\% depending on the patients risk factors. Patients with severe sepsis or septic shock are sicker, have longer hospital stays, are more frequently discharged to other short-term hospital or long-term care institutions, and represent the most expensive hospital condition treated in 20112.

The use of evidence-based practice (EBP) guidelines, such as the Surviving Sepsis Campaign (SSC), could lead to an earlier diagnosis, and consequently, earlier treatment. However, these guidelines have not been widely incorporated into clinical practice. The SSC is a compilation of international recommendations for the management of severe sepsis and shock. Many of these recommendations are interventions to prevent further system deterioration during and after diagnosis. Even when the presence of sepsis or progression to sepsis is suspected early in the course of treatment, timely implementation of adequate treatment management and guideline compliance are still a challenge. Therefore, the effectiveness of the guideline in preventing clinical complications for this population is still unclear to clinicians and researchers alike.

The majority of studies have focused on early detection and prevention of sepsis and little is known about the compliance rate to SSC and the impact of compliance on the prevention of sepsis-related complications. Further, the measurement of adherence to individual SSC recommendations rather than the entire SSC is, to our knowledge, limited. The majority of studies have used traditional randomized control trials with analytic techniques such as regression modeling to adjust for risk factors known from previous research. Data-driven methodologies, such as data mining techniques and machine learning, have the potential to identify new insights from electronic health records (EHRs) that can strengthen existing EBP guidelines.

The national mandate for all health professionals to implement interoperable EHRs by 2015 provides an opportunity for the reuse of potentially large amounts of EHR data to address new research questions that explore patterns of patient characteristics, evidence-based guideline interventions, and improvement in health. Furthermore, expanding the range of variables documented in EHRs to include team-based assessment and intervention data can increase our understanding of the compliance with EBP guidelines and the influence of these guidelines on patient outcomes. In the absence of such data elements, adherence to guidelines can only be inferred; it cannot be directly observed.

In this section, we present a methodology for using EHR data to estimate the compliance with the SSC guideline recommendations and also estimate the effect of the individual recommendations in the guideline on the prevention of in-hospital mortality and sepsis-related complications in patients with severe sepsis and septic shock.

\subsection{Methods}
Data from the EHR of a health system in the Midwest was transferred to a clinical data repository (CDR) at the University of Minnesota which is funded through a Clinical Translational Science Award. After IRB approval, de-identified data for all adult patients hospitalized between 1/1/09 to 12/31/11 with a severe sepsis or shock diagnosis was obtained for this study.

\subsubsection{Data and cohort selection}

The sample included 186 adult patients age 18 years or older with an ICD-9 diagnosis code of severe sepsis or shock (995.92 and 785.5*) identified from billing data. Since 785.* codes corresponding to shock can capture patients without sepsis, patients without severe sepsis or septic shock, and patients who did not receive antibiotics were excluded. These exclusions aimed to capture only those patients who had severe sepsis and septic shock, and were treated for that clinical condition. The final sample consisted of 177 patients.

\subsubsection{Variables of interest}

Fifteen predictor variables (baseline characteristics) were collected. These include sociodemographics and health disparities data: age, gender, race, ethnicity, and payer (Medicaid represents low income); laboratory results: lactate and white blood cells count (WBC); vital signs: heart rate (HR), respiratory rate (RR), temperature (Temp), mean arterial blood pressure (MAP); and diagnoses for respiratory, cardiovascular, cerebrovascular, and kidney-related comorbid conditions. ICD-9 codes for comorbid conditions were selected according to evidence in the literature . Comorbidities were aggregated from the patient’s prior problem list to detect preexisting (upon admission) respiratory, cardiovascular, cerebrovascular, and kidney problems. Each category was treated as yes/no if any of the ICD-9 codes in that category were present.

The outcomes of interest were inhospital mortality and development of new complications (respiratory, cardiovascular, cerebrovascular, and kidney) during the hospital encounter. New complications were determined as the presence of ICD-9 codes on the patients billing data that did not exist at the time of the admission.

\subsubsection{Study design}

This study aimed to analyze compliance with the SSC guideline recommendations in patients with severe sepsis or septic shock. Therefore, the baseline (TimeZero) was defined as the onset of sepsis and the patients were under observation until discharged. Unfortunately, the timestamp for the diagnoses is dated back to the time of admission; hence the onset of sepsis needs to be estimated. The onset time for sepsis was defined as the earliest time during a hospital encounter when the patient meets at least two of the following six criteria: MAP < 65, HR >100, RR >20, temperature < 95 or >100.94, WBC < 4 or > 12, and lactate > 2.0. The onset time was established based on current clinical practice and literature on sepsis5. The earliest time when two or more of these aforementioned conditions were met, a TimeZero flag was added to the time of first occurrence of that abnormality, and the timing of the SSC compliance commenced.

\subsubsection{Guideline compliance}

SSC guideline recommendations were translated into a readily computable set of rules. These rules have conditions related to an observation (e.g. MAP < 65 Hgmm) and an intervention to administer (e.g. give vasopressors) if the patient meets the condition of the rule. The SSC guideline was transformed into 15 rules in a computational format, one for each recommendation in the SSC guideline recommendations, and each rule was evaluated for each patient (see Figure 1). After each rule is an abbreviated name subsequently used in this paper.

\begin{figure}[ht!]
\centering
\includegraphics[width=70mm, height=100mm]{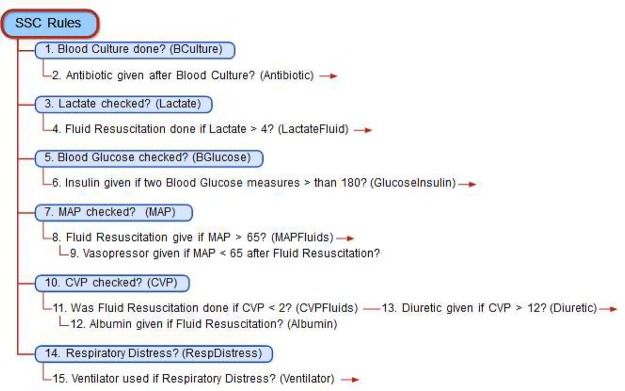}
\caption{SSC rules for measuring guideline compliances  \label{causa}}
\end{figure}

We call the treatment of a patient compliant (exposed) for a specific recommendation, if the patient meets the condition of the corresponding rule any time after TimeZero and the required intervention was administered; the treatment is non-compliant (unexposed) if the patient meets the condition of the corresponding rule after TimeZero, but the intervention was not administered (any time after TimeZero); and the recommendtion is not applicable to a treatment if the patient does not meet the condition of the corresponding rule. In estimating compliance (as a metric) with a specific recommendation, we simply measure the number of compliant encounters to which the recommendation is applicable. In this phase of the study, the time when a recommendation was administered was not incorporated in the analysis.

We also estimate the effect of the recommendation on the outcomes. We call a patient exposed to a recommendation, if the recommendation is applicable to the patient and the corresponding intervention was administered to the patient. We call a patient unexposed to a recommendation if the recommendation is applicable but was not applied (the treatment was non-compliant). The incidence fraction in exposed patients with respect to an outcome is the fraction of patients with the outcome among the exposed patients. The incidence fraction of the unexposed patients can be defined analogously. We define the effect of the recommendation on an outcome as the difference in the incidence fractions between the unexposed and exposed patients. The recommendation is beneficial (protective against an outcome) if the effect is positive, namely, the incidence faction in the unexposed is higher than the incidence fraction in the unexposed patients.

\subsubsection{Data quality}

Included variables were assessed for data quality regarding accuracy and completeness based on the literature and domain knowledge. Constraints were determined for plausible values, e.g., a CVP reading could not be greater than 50. Values outside of constraints were recoded as missing values. Any observation that took place before the estimated onset of sepsis (TimeZero) was considered a baseline observation. Simple mean imputation was the method of choice for imputing missing values. Imputation was necessary for lactate (7.7\%), temperature (3\%), and WBC (3\%). There was no missing data for the other variables and for the outcomes of interest. Central venous pressure was not included as a baseline characteristic due to the high number of missing values (54\%).

\subsubsection{Propensity score matching}

Patients who received SSC recommendations may be in worse health than patient who did not receive SSC recommendations. For example, patients whose lactate was measured may have more apparent (and possibly advanced) sepsis than patients whose lactate was not measured. To compensate for such disparities, propensity score matching (PSM) was employed. The goal of PSM is to balance the data set in terms of the covariates between patients exposed and unexposed to the SSC guideline recommendations. This is achieved by matching exposed patients with unexposed patients on their propensity (probability) of receiving the recommendations. This ensures that at TimeZero, pairs of patients, one exposed and one unexposed, are at the same state of health and they only differs in their exposure to the recommendation. PSM is a popular technique for estimating treatment effects.

To compute the propensity of patients to receive treatment, a logistic regression model was used, where the dependent variable is exposure to the recommendation and the independent variables are the covariates. The linear prediction (propensity score) of this model was computed for every patient. A new (matched) population was created from pairs of exposed and unexposed patients with matching propensity scores. Two scores match if they differ by no more than a certain caliper (.1 in our study). The effect of the recommendation was estimated by comparing the incident fraction among the exposed and unexposed patients in the matched population.

\subsubsection{PSM nested inside bootstrapping simulation}

In order to incorporate the effect of additional sources of variability arising due to estimation in the propensity score model and variability in the propensity score matched sample, 500 bootstrap samples were drawn from the original sample. In each of these bootstrap iterations, the propensity score model was estimated using the above caliper matching techniques and the effect of the recommendation was computed with respect to all outcomes. In recent years, bootstrap simulation has been widely employed in conjunction with PSM to better handle bias and confounding variables. For each recommendation and outcome, the 500 bootstrap iterations result in 500 estimates of the effect (of the recommendation on the outcome), approximating the sampling distribution of the effect.

\subsection{Results}
Table 1 shows the baseline characteristics of the study population. Results are reported as total count for categorical variables, and mean with inter-quartile (25\% to 75\%) range for continuous variables. As shown in Table 1, the majority of patients were male, Caucasian, and had Medicaid as the payer. Before the onset of sepsis, Cardiovascular comorbidities (56.4\%) were common, the mean HR (101.3) was slightly above the normal, as well as lactate (2.8), and WBC (15.8). The mean length of stay for the sample was 15 days, ranging from less than 24 hours to 6 months. TimeZero was within the first 24 hours of admission, and patients at that time were primarily (86.4\%) in the emergency department.

\begin{table}[ht]
\centering
\small
\begin{tabular}{lrr}
  \hline
Feature & Mean \\
\hline
Total Number of Patients &  177   \\ 
Average Age &  61  \\ 
Gender(Male) & 102 \\
Race(Caucasian) & 97 \\
Ethnicity(Latino) & 11 \\
Payer(Medicaid) & 102 \\
White Blood cell & 15.8 \\
Lactate & 28 \\
Mean blood Pressure & 73.9 \\
Temperature & 98.4 \\
Heart Rate & 101.3 \\
Respiratory Rate & 20.6 \\
Cardiovascular & 100 \\
Cerebrovascular & 66 \\
Respiratory & 69 \\
Kidney & 62 \\
\hline
\end{tabular}
\caption{Demographics statistics of patient population}\label{tbl:amstats}
\end{table}

In Figure 2, the effects of various rule-combination pairs are depicted. An effect is defined as the difference in the mean rate of progression to complications between the exposed and unexposed groups. Since we used bootstrap simulation, for each rule-complication pair, 500 replications were performed resulting in a sampling distribution for the effect. Sampling distribution for each rule-association pair is presented as boxplots. The boxplots represent the statistic measured, i.e. in this study, the differential impact of a recommendation on mortality between the exposed and unexposed population. When this statistic is 0, the recommendation has no effect. If the recommendation is greater than 0, it means that the recommendation is protective for that specific condition; and if the recommendation is below 0, the recommendation may even increase the risk for the outcome for that specific condition.

\begin{figure}[ht!]
\centering
\includegraphics[width=90mm, height=130mm]{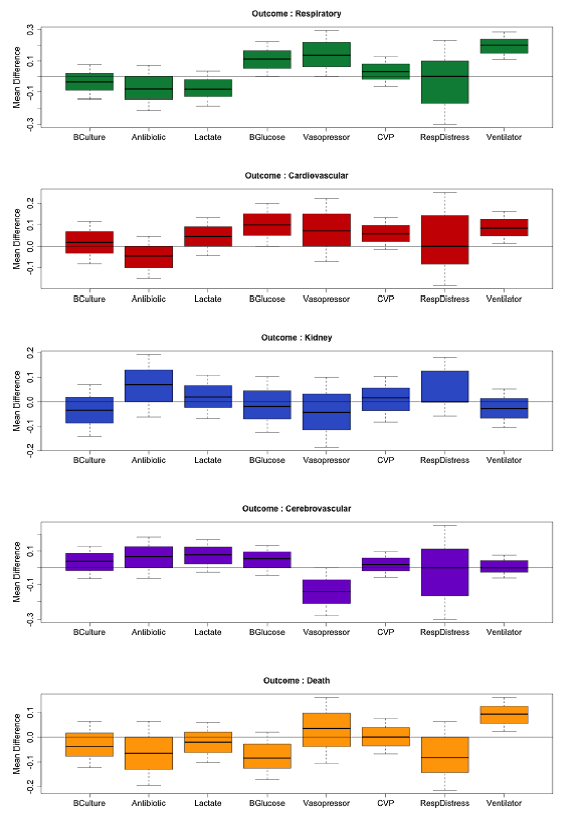}
\caption{Box-plots of the mean difference between groups (unexposed - exposed) to the guideline recommendations and each of the outcomes of interest.  \label{causa}}
\end{figure}

The panes (groups of boxplots) correspond to the complications and the boxes within each pane correspond to the recommendation (rule). For example, the effect of the Ventilator rule (Recommendation 15: patients in respiratory distress should be put on ventilator) on mortality (Death) is shown in the rightmost box (Ventilator) in the bottom-most pane (Death). Since all effects in the boxplot are above 0, namely the number of observed complications in the unexposed group is higher than in the exposed, compliance with the Ventilator rule reduces the number of deaths. Therefore, the corresponding recommendation is beneficial to protect patients from Death (mortality). In Table 3, we present the 95\% Confidence Intervals for various rule-outcome pairs.

95\% Confidence intervals for various rule-outcome pairs.
To further ensure the validity of the results, we examine the propensity score distribution in the exposed and unexposed group. As an example, Figure 3 illustrates the propensity score distribution for a randomly selected bootstrap iteration to measure the effect of Ventilator on Death. The horizontal axis represents the propensity score, which is the probability of receiving the interventions, and the vertical axis represents the density distribution, namely the proportion of patients in each group with a particular propensity for being put on Ventilator. Figure 3 shows substantial overlap between the propensity scores in the exposed and unexposed group. The propensity score overlap represents the distribution; the predictor Ventilator across the exposed and unexposed populations regarding the outcome Death; the balance was successful when the propensity score was applied for this population. Other rule-complication pairs exhibit similar propensity score distribution.

\begin{figure}[ht!]
\centering
\includegraphics[width=90mm, height=60mm]{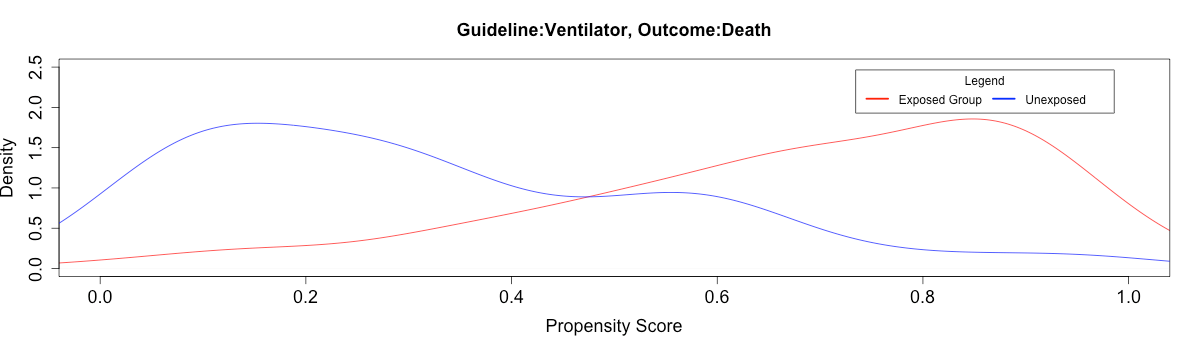}
\caption{Distribution of the propensity scores between exposed and unexposed groups for the outcome Death when patients and the SSC recommendation was Ventilator..  \label{causa}}
\end{figure}

\section{Conclusion}
The overall purpose of this study was to use EHR data to determine compliance with the Surviving Sepsis Campaign (SSC) guideline and measure its impact on inpatient mortality and sepsis complications in patients with severe sepsis and septic shock. Results showed that compliance with many of the recommendations was > 95\% for MAP and CVP with fluid resuscitation given for low readings. Other high compliance (greater than 80\%) recommendations were: insulin given for high blood glucose and evaluating respiratory distress. The recommendations with the lowest compliance (< 30\%) were: vasopressor or albumin for continuing low MAP or CVP readings. This may be due to a study design artifact, where the rule only considered interventions initiated after TimeZero (estimated onset of sepsis) while the fluid resuscitation may have taken place earlier. Alternatively, the apparently poor compliance could also be explained with issues related to the coding of fluids: during data validation, we found that it was difficult to track fluids.

Our study also demonstrates that retrospective EHR data can be used to evaluate the effect of compliance with guideline recommendations on outcomes. We found a number of SSC recommendations that were significantly protective against more than one complication: Ventilator was protective against Cardiovascular and Respiratory complications as well as Death; use of Vasopressors was protective for Respiratory complications.

Other recommendations, BCulture, Antibiotic, Vasopressor, Lactate, CVP, and RespDistress, showed results less consistent with our expectation. For instance, Vasopressor used to treat low MAP, appears to increase cerebrovascular complications. While this finding is not statistically significant, it may be congruent with the fact that small brain vessels are very sensitive to changes in blood pressure. Low MAP can cause oxygen deprivation, and consequently brain damage.

Ventilator, Vasopressor, and BGlucose showed protective effects against Respiratory complications. The SSC guideline recommends the implementation of ventilator therapy as soon as any change in respiratory status is noticed. This intervention aims to protect the patient against further system stress, restore hypoxia, help with perfusion across the main respiratory-cardio vessels, and decrease release of toxins due to respiratory efforts.

Our study is a proof-of-concept study demonstrating that EHR data can be used to estimate the effect of guideline recommendations. However, for several combinations of recommendations and outcomes, the effect was not significant. We believe that the reason is that guidelines represent workflows and the effect of the workflow goes beyond the effects of the individual guideline recommendations. For example, by considering the recommendations outside the context of the workflow, we may ignore whether the intervention addressed the condition that triggered its administration. If low MAP triggered the administration of vasopressors, without considering the workflow, we do not know whether MAP returned to the normal levels thereafter. Thus we cannot equate an adverse outcome with the failure of the guideline, it may be the result of the insufficiency of the intervention. Moving forward, we are going to model the workflows behind the guidelines and apply the same principles that we developed in this work to estimate the effect of the entire workflow.

This phase of our study did not address the timing of recommendations nor the time prior to TimeZero. For this analysis, guideline compliance was considered only after Time-Zero (the estimated onset), since compliance with SSC is only necessary in the presence of suspected or confirmed sepsis. There is no reason to suspect sepsis before TimeZero. However, some interventions may have started earlier, without respect to sepsis. For example, 100\% of the patients in this sample had antibiotics (potentially preventive antibiotics), but only 99 (55\%) patients received it after TimeZero.

The EHR does not provide date and time for certain ICD-9 diagnoses. During a hospital stay, all new diagnoses are recorded with the admission date. We know whether a diagnosis was present on admission or not, thus we know whether it is a preexisting or new condition, but do not know precisely when the patient developed this condition during the hospitalization. For this reason, we are unable to detect whether the SSC guideline was applied before or after a complication occurred, thus we may underestimate the beneficial effect of some of the recommendations. For example, high levels of lactate is highly related to hypoxia and pulmonary damage. If these patients were checked for lactate after pulmonary distress, we would consider the treatment compliant with the Lactate recommendation, but we would not know that the respiratory distress was already present at the time of the lactate measurement and we would incorrectly count it as a complication that the guideline failed to prevent.

\section{Complex Causal Rule Mining in  Irregular Time-Series Data}

\subsection{Introduction}

Effective management of human health remains a major societal challenge as evidenced by the rapid growth
in the number of patients with multiple chronic conditions. Type-II Diabetes Mellitus (T2DM), one of those conditions, affects 25.6 million (11.3\%) Americans of age 20 or older and is the seventh leading cause of death in the United States \cite{centers2011national}.  Effective treatment of T2DM is frequently complicated by diseases 
comorbid to T2DM, such as high blood pressure, high cholesterol, and abdominal obesity. Currently, these diseases are treated in isolation, which leads to wasteful duplicate treatments and suboptimal outcomes. The recent rise in the number of patients with multiple chronic conditions necessitates comprehensive treatment of these conditions to reduce medical waste and improve outcomes.

Finding optimal treatment for patients who suffer from multiple associated diseases, each of which can have
multiple available treatments is a complex problem.
%combinatorial pattern exploration problem. The quintessential tool for such problems is association rule mining. While association pattern mining has been successfully applied in many areas such as chemical informatics \cite{dehaspe1998finding}, sub-graph discovery \cite{chandola2009anomaly} and computer networks \cite{jiang2006research}, its adaption in health care research remains elusive. 
We could simply use techniques based on association, but a reasonable algorithm would likely find that 
the use of a drug is associated with some unfavorable outcome. This does not mean that the drug
is harmful; in fact in many cases, it simply means that patients who take the drug are sicker than those who do not and thus
they have a higher chance of the unfavorable outcome. What we really wish to know is whether a treatment \emph{causes} an unfavorable outcome, as opposed to being merely associated with it.

The difficulty in quantifying the effect of interventions on outcomes stems from subtle biases. 
Suppose we wish to quantify the effect of a cholesterol-lowering agent, statin, on diabetes.
We could simply compare the proportion of diabetic patients in the subpopulation that takes statin and the
subpopulation that does not and estimate the effect of statin as the difference between the two
proportions. This method would give the correct answer only if the statin-taking and non-statin-taking
patients are identical in all respects that influence the diabetes outcome. We refer to this situation as
treated and untreated patients being \emph{comparable}. Unfortunately, statin taking patients are not comparable to
non-statin-taking patients, because they take statin to treat high cholesterol, which by and in itself
increases the risk of diabetes. High cholesterol \emph{confounds} the effect of statin.
Many difference sources of bias exist, confounding is just one of the many. In this 
manuscript, we are going to address several different sources of bias, including confounding.

Techniques to address such biases in causal effect estimation exist.
However, these techniques have been designed to quantify the effect of a single intervention.
In trying to apply these techniques to our problem of finding optimal treatment 
for patients suffering from varying sets of diseases, we face two challenges.

First, patients with multiple conditions will likely need a combination of drugs.
Quantifying the effect of multiple concurrent interventions is semantically 
different from considering only a single intervention. 
The key concept in estimating the effect of an intervention is \emph{comparability}:
to estimate the effect of intervention, we need two groups of patients who are
identical in all relevant aspects except that one group receives the intervention
and the other group does not. For a single intervention, the first group is typically
the sickest patients who still do not get treated and the second group consists of
the healthiest patient who get treatment. They are reasonably in the same state of health.
However, when we go from a single intervention to multiple intervention and
try to estimate their \emph{joint} effect, comparability no longer exists.
A patient requiring multiple simultaneous interventions is so fundamentally different
from a patient who does not need any intervention that they are not comparable.

The other key challenge in finding optimal intervention sets for patients with
combinatorial sets of diseases is the combinatorial search space.
Even if we could trivially extend the methods for quantifying the effect of
a single intervention to a set of concurrent interventions, we would have to
systematically explore a combinatorially large search space.
The association rule mining framework \cite{agrawal1994fast} provides an efficient solution for
exploring combinatorial search spaces, however, it only detects associative
relationships. Our interest is in causal relationships.

In this manuscript, we propose causal rule mining, a framework for transitioning from association rule
mining towards causal inference in subpopulations. Specifically, given a set of interventions and a set of items to define subpopulations, we wish to find all subpopulations in which effective intervention combinations exist and in each such subpopulation, we wish to find all intervention combinations such that dropping any intervention from this combination will reduce the efficacy of the treatment.  We call these \emph{closed intervention sets}, which are not be confused with closed item sets.
As a concrete example, interventions can be drugs, subpopulations can be defined in terms of their
diseases and for each subpopulations (set of diseases), our algorithm would return effective drug 
cocktails of increasing number of constituent drugs. Leaving out any drug from the cocktail will reduce the
efficacy of the treatment. Closed intervention sets allow us to go from estimating a single intervention to
multiple interventions.

To address the exploration of the combinatorial search space, we propose a novel frequency-based
anti monotonic pruning strategy enable by the closed intervention set concept.
The essence of antimonotonic property is that if a set $I$ of interventions does not satisfy a criterion,
none of its supersets will. The proposed pruning strategy based on the closed intervention 
is strictly more efficient than the traditional pruning strategy used by the Apriori algorithm \cite{agrawal1994fast}.

Underneath our combinatorial exploration algorithm, we utilize the Rubin-Neyman model of causation \cite{sekhon2008neyman}. This model sets two conditions for causation: a set $X$ of interventions causes a change in $Y$ iff $X$ happens before $Y$ and $Y$ would be different had $X$ not occurred. 
The unobservable outcome of what would happen had a treated patient not received treatment
is a \emph{potential outcome} and needs to be estimated.
We present and compare five methods for estimating these potential outcomes and describe
the biases these methods can correct.

Typically the ground truth for the effect of drugs is not known. In order to assess the
quality of the estimates, we conduct a simulation study utilizing five different synthetic
data set that introduce a new source of bias. We will evaluate the effect of the bias
on the five proposed methods underscoring the statements with rigorous proofs when possible.

We also evaluate our work on a real clinical data set from Mayo Clinic. 
We have data for over 52,000 patients with 13 years of follow-up time. Our outcome
of interest is 5-year incident T2DM and we wish to extract patterns of interventions for patients suffering
from combinations of common comorbidities of T2DM.
First, we evaluate our methodology in terms of the computational cost, demonstrating the effectiveness
of the pruning methodologies. Next, we evaluate the patterns qualitatively, using patterns involving
statins. We show that our methodology extracted patterns that allow us to explain the controversial
patterns surrounding statin \cite{huupponen2013statins}. 

\myparag{Contributions.}
(1) We propose a novel framework for extracting 
causal rules from observational data correcting for a number of common biases. 
(2) We introduce the concept of closed intervention sets to extend the concept of 
quantifying the effect of a single intervention to a set of concurrent interventions
sidestepping the patient comparability problem.
Closed intervention sets also allow for a pruning strategy that is strictly more efficient than the traditional pruning strategy used by the Apriori algorithm \cite{agrawal1994fast}.
(3) We compare five methods of estimating causal effect from observational data
that are applicable to our problem and rigorously evaluate them on synthetic
data to mathematically prove (when possible) why they work.

\subsection{Background: Association Rule Mining}
We first briefly review the fundamental concepts of association rule mining
and extend these concepts to causal rule mining in the next section.
Consider a set $\mathcal{I}$ of \textbf{items}, which are single-term predicates evaluating to `true' or `false'.
For example, $\{age>55\}$ can be in item. A k-\textbf{itemset} is a set of $k$ items, evaluated as
the conjunction (logical 'and') of its constituent items.
Consider a dataset D = \{ $d_1,d_2.....d_n$ \}, which consists of $n$ \textbf{observations}. Each observation, denoted by $D_j$ is a set of items. %The items in the dataset can be denoted as \{ $x_1,x_2.....x_m$ \}.
An itemset $I={i_1,i_2,\ldots,i_k}$ ($I\subset\mathcal{I}$) \textbf{supports} an observation $D_j$ if
all items in $I$ evaluate to `true' in the observation.
The \textbf{support} of $I$ is the fraction of the observations in $D$ that support $I$.
An itemset is \textbf{frequent} if its support exceeds a pre-defined minimum support threshold.

A association rule is a logical implication of form $X \Rightarrow Y$, where $X$ and $Y$ are disjoint itemsets.
The support of a rule is $\support(XY)$ and the \textbf{confidence} of the rule is
\begin{displaymath}
\confidence(X\Rightarrow Y)=\frac{\support(XY)}{\support(X)}=\prob(Y|X).
\end{displaymath}

% --------------------------------------------------------------------------------------------------------------------------------------------
%                                                                                                                                                             Method
%                                                                                                                              Causal Rule Mining
%
\subsubsection{Causal Rule Mining}

Given an \textbf{intervention} itemset $X$ and an \textbf{outcome} item $Y$, such that
$X$ and $Y$ are disjoint, a causal rule is an implication of form $X \causes Y$,
suggesting that $X$ \emph{causes} a change in $Y$.
Let the itemset $S$ define a \textbf{subpopulation}, consisting of all observations that
support $S$. This subpopulation consists of all observations for which all items in $S$
evaluate to `true'.
The \textbf{causal rule} $X \causes Y|_S$ implies that the intervention $X$ has causal effect on $Y$
in the subpopulation defined by $S$.
The quantity of interest is the \textbf{causal effect}, which is the change in $Y$ in the subpopulation $S$
caused by $X$.  We will formally define the metric used to quantify the causal effect shortly.

%More generally, let $X=\{x_1, x_2, \ldots, x_k\}$ be an itemset, an \textbf{intervention set},
%and $Y$, an outcome.
%$X \causes Y$ is a causal rule, iff
%\begin{displaymath}
%\forall x\in X,\quad x \causes Y\;|_{X\setminus x}.
%\end{displaymath}
%An intervention set $X$ has to have significantly larger causal effect on $Y$
%than any of its subitemsets.

\myparag{Rubin-Neyman Causal Model.}
$X$ has a causal effect on $Y$ if (i) $X$ happens earlier than $Y$ and (ii) if $X$ had not happened, $Y$ would be different \cite{sekhon2008neyman}.

Our study design ensures that the intervention $X$ precedes the outcome $Y$,
but fulfilling the second conditions requires that we estimate the outcome for the same
patient both under intervention and without intervention.

\mysubparag{Potential Outcomes.}
Every patient in the dataset has two potential outcomes: $Y_0$ denotes their outcome had they not had the intervention $X$; and $Y_1$ denotes the outcome had they had the intervention. Typically, only one of the two potential outcomes can be observed. The observable outcome is the \textbf{actual} outcome (denoted by $Y$) and the unobservable potential outcome is called the \textbf{counterfactual} outcome.

Using the definition of counterfactual outcome, we can now define the metric for estimating the change in $Y$
caused by $X$. \textbf{Average Treatment response on the Treated} (ATT) is a widely known metric in the causal literature and is computed as follows:
\begin{equation*}
\att(X \causes Y|_S) = \E [ Y_1 - Y_0 ]_{X=1} = \E[Y_1]_{X=1}-\E[Y_0]_{X=1},
\end{equation*}
where $\E$ denotes the expectation and the $X=1$ in the subscript signals that we only evaluate the
expectation in the treated patients $(X=1)$.

ATT aims to compute an average per-patient change caused by the intervention. $Y_0$ = $Y_1$, indicates that the intervention resulted in no change in outcome for the patient.
%However, $Y_0$ > $Y_1$ or $Y_0$ < $Y_1$ are equally favorable responses and indicates that there is a measurable change brought upon by X on Y. \\

\begin{figure}[ht!]
\vspace{-3mm}
\centering
\includegraphics[width=60mm]{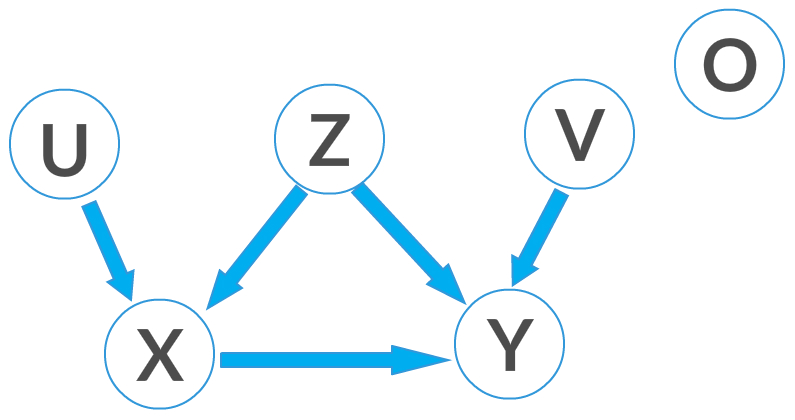}
\caption{Rubin-Neyman Causal Model \label{fig:causalGraph}}
\end{figure}

\mysubparag{Biases.}
Beside $X$, numerous other variables can also exert influence over $Y$, leading to biases in the
estimates.  To correct for these biases, we have correctly account for these other effects.
The quintessential tool for this purpose is the causal graph, depicted in Figure \ref{fig:causalGraph}.
The nodes of this graph are sets of variables that play a causal role and edges are
causal effects. This is not a correlation graph (or dependence graph),
because for example, $U$ and $Z$ are dependent given $X$, yet there is no edge between them.

Variables (items in $\mathcal{I}$) can exert influence on the effect of $X$ on $Y$ in three way: they may only
influence $X$, they may only influence $Y$ or them may influence both $X$ and $Y$.
Accordingly, variables can be categorized into four categories:
%\begin{description}
%\item[$V$] are variables that directly influence $Y$ and thus have \emph{direct effect} on Y; 
%\item[$U$] are variables that only influence $Y$ through $X$ and thus have \emph{indirect effect} on Y;
%\item[$Z$] are variables that influence both $X$ and $Y$ and are called \emph{confounders}; and finally
%\item[$O$]are variables that do not influence either $X$ or $Y$ and hence can be safely ignored.
%\end{description}

\begin{tabular}{cl}
$V$& are variables that directly influence $Y$ and thus have \\
  & \emph{direct effect} on $Y$ \\
$U$& are variables that only influence $Y$ through $X$ and \\
  &thus have \emph{indirect effect} on $Y$;\\
$Z$& are variables that influence both $X$ and $Y$ and are \\
  &called \emph{confounders}; and finally \\
$O$& are variables that do not influence either $X$ or $Y$ \\
  &and hence can be safely ignored.
\end{tabular}

%In order to estimate the causal effect of an intervention X on the outcome variable Y, while simultaneously accounting for confounding variables represented by Z, variables V, which affect only the outcome and variable U which affects the intervention and variables N which do not effect either the outcome and nor the intervention. e state-of-the-art metric used in estimating the effect of the intervention X is known as Average Treatment Effect on the treated (ATT). \\

Most of the causal inference literature assumes that the causal graph is known and true.
In other words, we know apriori which variables fall into each of the categories, $U$, $Z$,
$V$ and $O$. In our case, only $X$ and $Y$ are specified and we have to infer which category each other
variable (item) belongs to. Since this inference relies on association (dependence)
rather than causation, the discovered graph may have errors, misclassifications of variables
into the wrong category. For example, because of the marginal dependence between $U$ and $Y$,
variables in $U$ can easily get misclassified as $Z$. Such misclassifications do not necessarily
lead to biases, but they can cause loss of efficiency.

\myparag{Problem Formulation.}
Given a data set $D$, a set $\mathcal{S}$ of \textbf{subpopulation-defining} items,
a set $\mathcal{X}$ of \textbf{intervention} items, a minimal support threshold $\theta$
and a minimum effect threshold $\eta$, we wish to find all subpopulations $S$ ($S\subset\mathcal{S}$)
and all intervetions $X$ ($X\subset\mathcal{X}$), $X$ and $S$ are disjoint, such that
the causal rule $X\causes Y|_S$ is frequent and its intervention set $X$ is closed w.r.t. our metric of causal effect, ATT.

Note that the meaning of $\theta$, the minimum support threshold,
is different than in association rule mining literature. Typically, rules with support less than
$\theta$ are considered uninteresting, in other cases, it is simply a computational convenience,
but in our case, we set $\theta$ to a minimum value such that ATT is estimable for the discovered patterns.

We call a causal rule \textbf{frequent} iff its support exceeds the user-specified minimum threshold $\theta$
\begin{displaymath}
\support(X\causes Y|_S)=\support(XYS)=\prob(XYS) > \theta
\end{displaymath}
and we call an intervention set $X$ \textbf{closed} w.r.t. to ATT iff
\begin{displaymath}
\forall x\in X,\quad |ATT(x\causes Y|_{S,X\setminus x})| > \eta,
\end{displaymath}
where $\eta$ is the user-specified minimum causal effect threshold.
In other words, a causal rule is closed in a subpopulation, if its (absolute) effect is greater
than any of its sub-rules.

\mysubparag{Example.}
In a medical setting, $\mathcal{X}$ may be drugs, $\mathcal{S}$ could be comorbid diseases.
Then $X$ is a drug-combination that hopefully treats set of diseases $S$.
This set of drugs being \emph{closed} w.r.t. ATT means that dropping any drug from $X$ will
reduce the overall efficacy of the treatment; the patient is not taking unnecessary drugs.

An itemset is closed if its support is strictly higher than all of its subitemsets'. Analogously, an
intervention set is closed if its absolute causal effect is strictly higher than all of its subitemsets'.

% --------------------------------------------------------------------------------------------------------------------------------------------
%                                                                                                                                                             Method
%                                                                                                                                            Algorithm
%
\subsubsection{Frequent Causal Pattern Mining Algorithm}

We can now present our algorithm for causal pattern mining. At a very high level, the algorithm comprises of two nested frequent pattern enumeration \cite{goethals2003survey} loops. The outer loop enumerates subpopulation-defining itemsets $S$ using items in $\mathcal{S}$, while the inner loop enumerates intervention combinations using items in $\mathcal{X}\setminus\mathcal{S}$. More generally, $\mathcal{X}$ and $\mathcal{S}$ can overlap but we do not consider that in this paper. 
Effective algorithms to this end exists \cite{han2000freespan,han2004mining}, we simply use Apriori \cite{agrawal1994fast}.

Once the patterns are discovered, the ATT of the interventions are computed, using one of the methods
from Section \ref{sec:CausalMetrics} and the frequent, effective patterns are returned.

On the surface, this approach appears very expensive, however several novel, extremely effective pruning strategies are possible and we describe them below.

\myparag{Potential Outcome Support Pruning.}
Let $X$ be an intervention $k$-itemset, $S$ be a subpopulation-defining itemset, and let $X$ and $S$ be disjoint.
Further, $X_{-i}$ be an itemset that evaluates to `true' iff all items except the $i$th are `true' but the $i$th item is `false'. Using association rule mining terminology, all items in $X$ except the $i$th are present in the transaction.

\begin{defn}[Potential Outcome Support Pruning]
We only need to consider itemsets $X$ such that
\begin{eqnarray*}
\min \{ \support(S,X), & \support(\{S,X_{-1}), \ldots, &\\
&\quad\support(S,X_{-k}) \} & > \theta.
\end{eqnarray*}
\end{defn}

In order to be able to estimate the effect of $x\in X$ in the subpopulation $S$, we need to have
observations with $x$ `true' and also with $x$ `false' in $S$.

\begin{lemma}
Potential Outcome Support Pruning is antimonotonic.
\end{lemma}

\noindent\textsc{Proof:}
Consider a causal rule $X\causes Y|_S$. If the causal rule $X\causes Y|_S$ is infrequent, then
\begin{displaymath}
\support(XS) < \theta \quad\lor\quad \exists i, \support(X_{-i}S) < \theta.
\end{displaymath}
If $\support(X_{-i}S)$ had insufficient support, then any extension of it with an intervention item
$x$ will continue to have insufficient support, thus the $Xx\causes Y|_S$ rule will have
insufficient support. Likewise, if $\support(XS)$ had insufficient support, then any extension of it with an intervention item $x$
will also have insufficient support.

\myparag{Pruning based on Causal Effect.}

\begin{proposition}
Effective causal rule pruning condition is anti-monotonic.
\end{proposition}

\textsc{Rationale:}
To explain the rational, let us return to the medical example, where $X$ is a combination of
drugs forming a treatment.  Assuming that the effects of drugs are additive,
if a casual rule $X\causes Y|_S$ is ineffective because
\begin{displaymath}
\exists x_i\in X,\qquad |\att(x_i\causes Y|_{S,X\setminus x_i}) | < \eta,
\end{displaymath}
then forming a new rule $Xx_j\causes Y|_S$ will also be ineffective because
\begin{displaymath}
|\att(x_i \causes Y|_{S,x_j,X\setminus x_i}) |
\end{displaymath}
will be ineffective.  In the presence of positive interactions (that reinforce each other's effect)
among the drugs, this statement may not hold true.
Beside statistical reasoning, one can question why a patient should receive a drug that has no effect
in a combination.

%
%\textbf{Lemma 3:} Causal pattern mining framework is complete. \\
%\textbf{Proof:} Causal pattern mining framework is able to extract all causal rules of the form $X\causes Y|_S$, such that the rules are effective and frequent. Our framework is complete since for a given subpopulation named S and a given outcome denoted by Y, we identify all possible interventions called X,  in an apriori fashion. \\
%
%\textbf{Lemma 4:}  Causal pattern mining framework is correct. \\
%\textbf{Proof:} Causal pattern mining framework is able to extract all causal rules of the form $X\causes Y|_S$. The correctness of these rules depends upon the chosen metric for ATT estimation. In the following section, we would be discussing the metrics in greater detail. \\

% --------------------------------------------------------------------------------------------------------------------------------------------
%                                                                                                                                                             Method
%                                                                                                                        Causal Estim Methods
%
\subsection{Causal Estimation Methods}\label{sec:CausalMetrics}
%In order to estimate $ATT(X \causes Y|_S)$ we need to compute $\E[Y_1]$ and $\E[Y_0]$. The challenge then lies in the evaluation of $\E[Y_0]$ as it represents the average counterfactual outcome, had the patients not prescribed the intervention. Performing an Randomized control trial \cite{matthews2006introduction} is one of the ways to estimate E($Y_0$). RCT's have been widely employed in the past for estimating the effects of interventions \cite{oldridge1988cardiac},  \cite{rahimtoola1985perspective}. However RCT's are expensive to perform as as they often require substantially large patient populations for estimating the effect of interventions. To overcome this challenge, in this section, we would be discussing various metrics for estimating the counterfactual outcome $Y_0$. \\

ATT, our metric of interest, with respect to a single intervention $x$ in a subpopulation $S$ is defined as
\begin{displaymath}
\att(x\causes Y|_S) =\E\left[Y_1 - Y_0\right]_{S,X=1},
\end{displaymath}
which is the expected difference between the potential outcome under treatment $Y_1$ and the potential outcome
without treatment $Y_0$ in patients with $S$ who actually received treatment.
Since we consider treated patients, the potential outcome $Y_1$ can be observed, the potential outcome $Y_0$
cannot. Thus at least one of the two must be estimated.
The methods we present below differ in which potential outcome they estimate and how they estimate it.

For the discussion below, we consider the variables $X$, $Z$, $U$ and $V$ from the causal graph in Figure \ref{fig:causalGraph}. $X$ is a single intervention, $U$, $V$ and $Z$ can be sets of items.
For regression models, we will denote the matrix defined by $U$, $V$ and $Z$ in the subpopulation $S$
as $U$, $V$ and $Z$ (same letter as the variable sets).

\myparag{Counterfactual Confidence (CC).}
This is the simplest method. We simply assume that the patients who receive intervention $X=1$ and
those who do not $X=0$, do not differ in any important respect that would influence $Y$.
Under this assumption, $Y_1$ in the treated is simply the actual outcome in the treated and the
potential outcome $Y_0$ is simply the actual outcome in the non-treated ($X=0$).
Thus
 \begin{eqnarray*}
\att &=&  \confidence((X=1) \causes Y|_S) -
\confidence((X=0) \causes Y|_S), \\
&=&\prob(Y|S, X=1)-\prob(Y|S, X=0)
%\end{flalign}
\end{eqnarray*}

In the followings, to improve readability, we drop the $S$ subscript.  All evaluations take place
in the $S$ subpopulations.

\myparag{Direct Adjustment (DA). }
We cannot estimate $Y_0$ in the treated ($X=1$) as the actual outcome $Y$ in the untreated, because
the treated and untreated populations can significantly differ in variables such as $Z$ and $V$ that
influence $Y$.  In Direct Adjustment, we attempt to directly remove the effect of $V$ and $Z$
by including them into a regression model.  Since a regression model relates the means of the predictors
with the mean of the outcome, we can remove the effect of $V$ and $Z$ by making their means 0.

Let $R$ be a generalized linear regression model, predicting $Y$ via a link function $g$
\begin{displaymath}
g(Y|V, Z, X) = \beta_0+\beta_VV+\beta_ZZ+\beta_XX.
\end{displaymath}
Then the (link-transformed) potential outcome under treatment is
$g(Y_1)=\beta0+\beta_VV+\beta_ZZ+\beta_X$ and the potential
outcome without treatment is $g(Y_0)=\beta0+\beta_VV+\beta_ZZ$.
The ATT is then
\begin{eqnarray*}
\att&=&\E\left[g^{-1}(Y_1|V,Z,X=1)\right]_{X=1} -\\
&&\qquad \E\left[g^{-1}(Y_0|V,Z,X=0)\right]_{X=1}.
\end{eqnarray*}
where $g^{-1}(Y_1|V,Z,X=1)$ is prediction for an observation with the observed $V$ and $Z$ but
with $X$ set to 1.  The $\E(\cdot)_{X=1}$ notation signifies that these expectation of the predictions are
taken only over patients who actually received the treatment.

The advantage of DA (over CC) is manyfold. First, it can adjust for $Z$ and $V$ as long the model
specification is correct, namely the interaction terms that may exist among $Z$ and $V$ are specified correctly.
Second, we get correct estimates even if we ignore $U$, because $U$ is conditionally independent of $Y$
given $X$. This unfortunately only is a theoretical advantage, because we have to infer from the data whether a
variable is a predictor of $Y$ and $U$ is marginally dependent on $Y$, so we will likely adjust for $U$, even if we don't need to.

\myparag{Counterfactual Model (CM).}
In this technique, we build an explicit model for the potential outcome without treatment $Y_0$
using patients with $X=0$.  Specifically, we build a model
\begin{equation*}
g(Y|V,Z,X=0)=\beta_0+\beta_VV+\beta_ZZ.
\end{equation*}
and estimate the potential outcome as
\begin{equation*}
g(Y_0|V,Z)=g(Y|V,Z,X=0).
\end{equation*}
The ATT is then
\begin{equation*}
\att = \prob(Y|X=1) - \E\left[ g^{-1}(Y_0|V,Z)\right]_{X=1}.
\end{equation*}

Similarly to Direct Adjustment, the Counterfactual Model does not depend on $U$. However,
in case of the Counterfactual Model, we are only considering the population with $X=0$. In
this population, $U$ and $Y$ are independent, thus we will not include $U$ variables into the model.

\myparag{Propensity Score Matching (PSM).}
The central idea of Propensity Score Matching is to create a new population, such that patients
in this new population are comparable in all relevant respects and thus the expectation of the potential outcome
in the untreated equals the expectation of the actual outcome in the untreated.

Patients are matched based on their propensity of receiving treatment. This propensity is computed as
a logistic regression model with treatment as the dependent variable
\begin{displaymath}
\log\frac{\prob(X)}{1-\prob(X)}=\beta_0 + \beta_VV+\beta_ZZ.
\end{displaymath}
Patient pairs are formed, such that in each pair, one patient received treatment and the other did not
and their propensities for treatment differ by no more than a user-defined caliper difference $\rho$.

The matched population has an equal number of treated and untreated patients, is balanced on $V$ and $Z$,
thus the patients are comparable in terms of their baseline risk of $Y$. Hopefully, the only factor causing a difference
in outcome is the treatment.

For estimating ATT, the potential outcome without treatment is estimated from the actual outcomes
of the patients in the matched population who did not receive treatment:
\begin{eqnarray*}
ATT &=& \E\left[ Y_1 - Y_0 \right] \\
&-& \prob(Y|X=1,M)-\prob(Y|X=0,M),
\end{eqnarray*}
where M denotes the matched population.

Among the methods we consider, propensity score matching most strictly enforces the patient comparability criterion, 
however, it is susceptible to misspecification of the propensity regression model, which
can erode the quality of the matching.

 \myparag{Stratified Non-Parametric (SN).}
In the stratified estimation, we directly compute the expectation via stratification. The assumption is that
 the patient in each stratum are comparable in all relevant respects and only differ in the presence or
 absence of intervention. In each stratum, we can estimate the potential outcome $Y_0$ in the treated
 as the actual outcome $Y$ in the untreated.

%In this technique, the causal estimate $ATT(X \causes Y|_S)$ is computed in a non-parametric way. Further, this technique is robust to model-misspecification. \\
%
% $N_{v}$ : Total number of potential categorical values or combinations for V. \\
%
% $N_{u}$ : Total number of potential categorical values or combinations for U. \\
%
% $N_{z}$ : Total number of potential categorical values or combinations for Z. \\
%
%
%\begin{displaymath}
%ATT(X->Y)  = \sum_{v=1}^{N_v}  \sum_{u=1}^{N_u} \sum_{z=1}^{N_z}  || P(Y/X=1) - P(Y/X=0) ||
% \end{displaymath}

 \begin{eqnarray*}
 ATT &=& \E\left[Y_1-Y_0\right]_{X=1} \\
 &=& \sum_l P(l|X=1) \left[ P(Y_1|l,X=1)-P(Y_0|l,X=1) \right] \\
 &=& \sum_l P(l|X=1) \left[ P(Y|X=1)-P(Y|X=0) \right],
\end{eqnarray*}
where $l$ iterates over the combined levels of $V$ and $Z$.
If we can identify the items that fall into $U$, then we can ignore them, otherwise, we should
include them as well into the stratification.

The stratified method makes very few assumptions and should arrive at the correct estimate
as long as each of the strata are sufficiently large. The key disadvantage of the stratified method
lies in stratification itself: when the number of items across which we need to stratify is too large,
we may end up dividing the population into excessively many small subpopulations (strata) and
become unable to estimate the causal effect in many of them thus introducing bias into the estimate.
%
%The causal estimate  is computed across every possible value of U,V and Z.  The effect of $X$ is heterogeneous across the levels of $Z, V$. \\

% --------------------------------------------------------------------------------------------------------------------------------------------
%                                                                                                                                                             Results
%

\subsection{Results}

After describing our data and study design, we present three evaluations of the proposed methodology.
The first evaluation demonstrates the computational efficiency of our pruning methodologies, isolating the
effect of each pruning methods: (i) Apriori support-based pruning, (ii) Potential Outcome Support Pruning, and (iii)
Potential Outcome Support Pruning in conjunction with Effective Causal Rule Pruning.
In the second section, we provide a qualitative evaluation, looking at patterns involving statin.
We attempt to use the extracted patterns to explain the controversial findings that exist in the literature
regarding the effect of statin on diabetes.
Finally, in order to compare the treatment effect estimates to a ground truth, which does not exits for real drugs,
we simulate a data set using proportions we derived from the Mayo Clinic data set.

\myparag{Data and Study Design.}
In this study we utilized a large cohort of Mayo Clinic patients with  data between 1999 and 2013.  We included all adult patients (69,747) with research consent. The baseline of our study was set at Jan. 1, 2005. 
We collected lab results, medications, vital signs and status, and medication orders during a 6-year \emph{retrospective period}
between 1999 and the baseline to ascertain the patient's baseline comorbidities.
From this cohort, we excluded all patients with a diagnosis of diabetes before the baseline (478 patients), missing fasting plasma glucose measurements (14,559 patients), patients whose lipid health could not be determined (1,023 patients) and patients with unknown hypertension status (498 patients). Our final study cohort consists of 52,139 patients who were followed until the summer of 2013. 

Patients were phenotyped during the retrospective period. Comorbidities of interest include Impaired Fasting
Glucose (IFG), abdominal obesity, Hypertension (HTN; high blood pressure) and hyperlipidemia (HLP; high cholesterol).  For each comorbidity, the phenotyping algorithm classified patients into three broad levels
of severity: normal, mild and severe. Normal patients show no sign of disease; mild patients are either
untreated and out of control or are controlled using first-line therapy; severe patients require more
aggressive therapy.  IFG is categorized into normal and pre-diabetic, the latter indicating impaired fasting
plasma glucose levels but not meeting the diabetes criteria yet. For this study, progression to T2DM within 5 years from baseline (i.e. Jan 1, 2005) was chosen as our outcome of interest. Out of 52,139 patients 3627 patients progressed to T2DM , 41028 patients did not progressed to T2DM and the remaining patients (7484) dropped out of the study. In Table \ref{tbl:stats} we present 
statistics about our patient population.
%
%\subsubsection{DataTable}
%
%In Table \ref{tbl:stats} we present the patient population statistics. Apart from demographic information such as age and gender, we also present health information such as patients Impaired Fasting Glucose (IFG), Obesity, Hypertension and Hyperlipidemia. Obesity, Hypertension and Hyperlipidemia are diseases often associated with T2DM. For preventing and managing such diseases patients are often prescribed drugs such as statin, vibrates, cholesterol, acerb, diuret and beta blockers. All the attributes are binary in nature. For example age has been discretized as Age less than 45, Age greater than 45 and less than 55 and Age greater than 55 respectively. \\
 
\begin{table}[ht]
\centering
\small
\begin{tabular}{lrr}
  \hline
 &\multicolumn{2}{c}{T2DM}\\
 \cline{2-3}
& Present & Absent  \\ 
\hline
Total Number of Patients &  3627 & 41028   \\ 
Average Age &  44.73  &  35.58  \\ 
Male(\%)  &   51 & 41 \\
Female(\%)  & 49   &  59 \\
\hline
\multicolumn{3}{c}{Patient Diagnosis Status (\%)}\\
NormFG & 42 & 84 \\
PreDM & 58 & 16  \\
Normal Obesity & 29& 59 \\
Mild Obesity & 25 & 30\\
Severe Obesity & 46 & 11\\
Normal Hypertension & 48& 69\\
Mild Hypertension & 33 & 23\\
Severe Hypertension & 19  & 08  \\
Normal Hyperlipidemia & 12 & 29\\
Mild Hyperlipidemia & 72 & 64\\
Severe Hyperlipidemia & 16 & 07\\
\hline
\multicolumn{3}{c}{Patient Medication Status(\%)}\\
Statin & 26 & 11\\
Fibrates & 03& 01\\
Cholesterol.Other & 02 & 01\\
Acerab & 17 & 07\\
Diuret & 18& 07\\
CCB & 08 & 04\\
BetaBlockers & 22 & 10 \\
HTN.Others & 01 & 01\\
\hline
\end{tabular}
\caption{Demographics statistics of patient population}\label{tbl:stats}
\end{table}

% --------------------------------------------------------------------------------------------------------------------------------------------
%                                                                                                                                                            Results
%                                                                                                                              Pruning Efficiency
%
\subsubsection{Pruning Efficiency}

In our work, we proposed two new pruning methods. First, we have the Potential Outcome Support
Pruning, which aims to eliminate patterns for which the ATT is not estimable. Second, we have
the Effective Causal Rule Pruning, where we eliminate patterns that do not improve treatment effectiveness
relative to the subitemsets. 

In Figure \ref{causa} we present the number of patterns discovered using (i) the traditional Apriori support based pruning, (ii) our proposed Potential Outcome Support Pruning (POSP), and (iii) POSP in conjunction with Effective Causal Rule Pruning (ECRP).  

\begin{figure}[ht!]
\centering
\includegraphics[width=60mm, height=50mm]{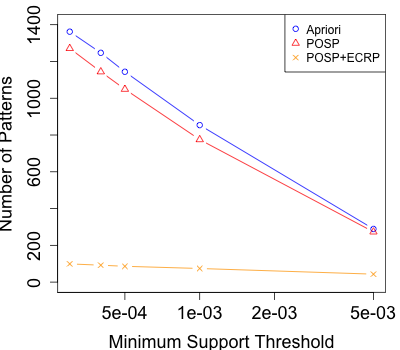}
\caption{Comparison of Pruning Techniques  \label{causa}}
\end{figure}

The number of patterns discovered by POSP is strictly less than the number of patterns discovered by
the Apriori pruning. POSP in conjunction with ECRP is very effective. 

% --------------------------------------------------------------------------------------------------------------------------------------------
%                                                                                                                                                            Results
%                                                                                                                                                  Statin
%
\subsubsection{Statin}

In this section, we demonstrate that the proposed causal rule mining methodology can be used to
discover non-trivial patterns from the above diabetes data set.

In recent years, the use of statins, a class of cholesterol-lowering agents, have been prescribed
increasingly. High cholesterol (hyperlipidemia) is linked to cardio-vascular mortality and the efficacy
of statins in reducing cardio-vascular mortality is well documented. However, as evidenced by a 2013
BMJ editorial \cite{huupponen2013statins} devoted to this topic, statins are surrounded in controversy. In patients with normal
blood sugar levels (labeled as NormalFG), statins have a detrimental effect, they increase the risk
of diabetes; yet in pre-diabetic patients (PreDM), it appears to have no effect.
What we demonstrate below is that this phenomenon is simply disease heterogeneity.

First, we describe how this problem maps to the causal rule mining problem.
Our set of interventions ($\mathcal{X}$) consists of statin and our subpopulation
defining variables consist of the various levels of HTN, HLP and IFG ($\mathcal{S}$).
Our interest is the effect of statin ($x$) on T2DM ($Y$)
in all possible subpopulations $S$, $S\subset\mathcal{S}$.

In this setup, HTN, which is associated with both hyperlipidemia (and statin use), as well as with
T2DM, is a confounder ($Z$). A cholesterol drug, other than statin, (say) fibrates, are in the
$U$ category: they are predictive of statin (patients on monotherapy who take fibrates do not
take statins), but have no effect on $Y$, because its effect is already incorporated into the
hyperlipidemia severity variables that defined the subpopulation.
Variables that only influence diabetes but not statin use (say a diabetes drug) would fall into
the $V$ category.
All subpopulations have variables that fall into $Z$ and $U$ and some subpopulation may also
have $V$.

The HLP variable in Table \ref{tbl:stats} uses statin as part of its definition, thus we constructed
two new variables.  The first one is HLP1, a variable at the borderline between HLP-Normal
and HLP-Mild, consisting of untreated patients with mildly abnormal lab results (these fall into HLP-Normal) 
and patients who are diagnosed and receive a first-line treatment (they fall into HLP-Mild). 
Comparability is the central concept of estimating causal effects and these patients are comparable at baseline.
Similarly, we also created another variable, HLP2, which is at the border of HLP-Mild and HLP-Severe, 
again consisting of patients who are comparable in relevant aspects of their health at baseline.

% latex table generated in R 3.2.0 by xtable 1.8-0 package
% Wed Feb 10 20:14:46 2016
\begin{table}[ht]
\centering
\begin{tabular}{lrrrrr}
  \hline
$S$ & CC & DA & CM & PSM & SN \\ 
  \hline
PreDM & 0.145 & 0.022 & 0.010 & 0.022 & 0.017 \\ 
  NormFG & 0.060 & 0.023 & 0.034 & 0.017 & 0.029 \\ 
  HLP1 & 0.078 & 0.019 & 0.014 & 0.010 & 0.010 \\ 
  HLP2 & 0.021 & -0.013 & -0.010 & -0.021 & -0.015 \\ 
  PreDM,HLP1 & 0.067 & 0.018 & 0.021 & 0.004 & 0.002 \\ 
 PreDM,HLP2 & 0.001 & -0.038 & -0.031 & -0.048 & -0.043 \\ 
  NormFG,HLP1 & 0.043 & 0.020 & 0.015 & 0.014 & 0.013 \\ 
  NormFG,HLP2 & 0.017 & -0.002 & -0.002 & -0.005 & -0.004 \\ 
   \hline
\end{tabular}
\caption{ATT due to statin in various subpopulations $S$ as estimated by
the 5 proposed methods.}\label{tbl:statin}
\end{table}

Table \ref{tbl:statin} presents the ATT estimates obtained by the various methods
proposed in Section 3.4 for some of the most relevant subpopulations.
Negative ATT indicates beneficial effect and positive ATT indicates detrimental effect.

Counterfactual confidence (CC) estimates statin to be detrimental in all subpopulations.
While statins are known to have detrimental effect in patients with normal glucose levels \cite{huupponen2013statins},
it is unlikely that statins are universally detrimental, even in patients with severe hyperlipidemia,
the very disease it is supposed to treat.

The results between DA, CM, PSM and SN are similar, with PSM and SN having larger
effect sizes in general. The picture that emerges from these results is that patients
with severe hyperlipidemia appear to benefit from statin treatment even in terms of
their diabetes outcomes, while statin treatment is moderately detrimental for patients
with mild hyperlipidemia.

Bootstrap estimation was used to compute the statistical significance of these results.
For brevity, we report the results only for PSM. 
The estimates are significant in the following subpopulations: NormFG, PreDM+HLP2
(p-values are <.001) and NormFG+HLP1 (p-value .05).

The true ATT in these subpopulations is not know.  To investigate the accuracy that the various
methods achieve, we use simulated that is largely based on this example \cite{huupponen2013statins, castrostatin}.

% --------------------------------------------------------------------------------------------------------------------------------------------
%                                                                                                                                                            Results
%                                                                                                                              Synthetic Data
%
\subsubsection{Synthetic Data}
In this section, we describe four experiments utilizing synthetic data sets, 
each of which introduces a new potential source of bias.
Our objective is to illustrate the ability of the five methods from Section 3.4
for adjusting for these biases. We compare their ATT estimates to the true ATT we used 
to generate the data set and discuss reasons for their success or failure.\\
The rows of Table \ref{tbl:synthetic} correspond to the synthetic data sets in increasing order of
the biases we introduced and the columns corresponds to the methods: Conf (confidence), 
CC (Counterfactual Confidence), DA (Direct Adjustment), CM (Counterfactual Model), PSM
(Propensity Score Matching) and SNP (Stratified Non-Parametric). \\
Some of these methods, DA, CM, PSM and SNP take the causal graph structure into account while estimating ATT.
Specifically, they require the information whether a variable is a confounder ($Z$), has a direct effect ($V$),
an indirect effect ($V$), or no effect ($O$). 
%This information is not known, it needs to be inferred
%from the data. When the inferred causal graph differs from the true causal graph, the table
%has two columns for the method: one showing tis performance under the true graph (`True') 
%and one under the discovered (`Disc') graph. 
PSM and SNP are not sensitive to the inclusion
of superfluous variables, they simply decrease the method's efficiency. \\
In all of the data sets, we use a notation consistent with Figure 1: $Z$ is the central
disease with outcome $Y$; $X$ is the intervention of interest that treats $Z$; $V$ is another 
disease with direct causal effect on $Y$, but $V$ is independent of $X$; and $U$ is a third disease,
which can be treated with $X$, but has no impact on $Y$.
All data sets contain 5000 observations.

\experiment{I. Direct Causal Effect from $V$.}
We assume that every patient in the cohort has disease $Z$ at the same severity. They
are all comparable w.r.t. $Z$.
30\% of the patients are subject to the intervention $X$ aimed at treating $Z$, while others are not.
Untreated patients face a 25\% chance of having $Y$, while treated patients only have 10\% chance.
Some patients, 20\% of the population, also have disease $V$, which directly affects $Y$: 
it increases the probability of $Y$ by 5\%. \\
In this example the true ATT is -.15, as $X$ reduces the chance of $Y$ by 15\%.
Our causal graph dictates that $X$ and $V$ be marginally independent, hence this
this effect is homogeneous across the levels of $V$. (Otherwise $V$ would become
predictive of $X$ and it would become a confounder. Confounding is discussed in experiments III-V.)
All methods estimated the ATT correctly, because ATT does not depend on $V$.
We can demonstrate this by stratifying on $V$ and using the marginal independence of $X$ and $V$.
\begin{eqnarray*}\hspace{-4mm}
ATT &=& \E\left[ \prob(Y|X=1) - \prob(Y|X=0) \right] \\
       &=& \sum_{v\in V} \prob(V=v) \left[ \prob(Y|V=v,X=1)-\prob(Y|V=v,X=0) \right] \\
       &=& \sum_{v\in V} \left[ \prob(Y,V=v|X=1)-\prob(Y,V=v|X=0) \right] \\
       &=& \prob(Y|X=1) - \prob(Y|X=0)
\end{eqnarray*}
where $v$ denotes the levels of $V$.  The marginal independence of $X$ and $V$ is
used in step three:
\begin{displaymath}
\prob(Y|V,X)=\frac{\prob(Y,V,X)}{\prob(V,X)}=\frac{\prob(Y,V|X)\prob(X)}{\prob(X,V)}=\frac{\prob(Y,V|X)}{\prob(V)}.
\end{displaymath}

\experiment{II. Indirect Causal Effect.}
The setup for this experiment is the same as for the 'Direct Causal Effect' experiment, 
except we have disease $U$ instead of $V$. Just like $Z$, disease $U$ is also treated by $X$,
but $U$ has no direct effect on $Y$; its effect is indirect through $X$. $U$ is thus independent of
$Y$ given $X$. The true ATT continues to be -.15. \\
Again, the ATT does not depend on $U$, hence all methods estimated it correctly.
To demonstrate that ATT does not depend on $U$, we use stratification and the conditional
independence of $Y$ and $U$.
\begin{eqnarray*}
ATT &=& \E\left[ \prob(Y|X=1)-\prob(Y|X=0) \right] \\
&=& \sum_{u\in U} \left[ \prob(Y|U=u,X=1)\prob(U=u|X=1) \right. \\
&&\qquad \left. - \prob(Y|U=u,X=0)\prob(U=u|X=0) \right] \\
&=& \sum_{u\in U} \left[ \prob(Y|X=1)\prob(U=u|X=1) \right. \\
&& \qquad \left.- \prob(Y|X=0)\prob(U=u|X=0) \right] \\ 
&=& \prob(Y|X=1)\sum_u\prob(U=u|X=1) - \\
&&\qquad \prob(Y|X=0)\sum_u\prob(U=u|X=0) \\
&=& \prob(Y|X=1)-\prob(Y|X=0)
\end{eqnarray*}

\experiment{III. Confounding.}
In this experiment, we consider the simplest case of confounding, involving a single disease $Z$,
a single treatment $X$ and outcome $Y$.
20\% of the patients have disease $Z$ and 95\% of the diseased patients are treated with $X$,
while 5\% are not. All treated patients have $Z$.
25\% of the untreated patients ($Z=1$ and $X=0$) have outcome $Y$; 10\% of the treated
patients ($Z=1$ and $X=1$) have the outcome; and only 5\% of the healthy patients ($Z=0$) have it.
The true ATT is -.15. \\
In the presence of confounding, the counterfactual confidence and ATT do not coincide.
With $z$ denoting the levels of $Z$ and $\prob(z)$ being a shorthand for $\prob(Z=z)$,
\begin{eqnarray*}
ATT &=& \E \left[ \prob(Y|X=1)-\prob(Y|X=0)\right] \\
&=&\sum_z \prob(z) \left[ \prob(Y|X=1,z) - \prob(Y|X=0,z) \right],
\end{eqnarray*}
while the counterfactual confidence (CC) is 
\begin{eqnarray*}
CC &=& \prob(Y|X=1)-\prob(Y|X=0)\\
&=& \sum_z \left[ \prob(Y|X=1,z)\prob(z|X=1) \right.\\
&&\qquad \left. - \prob(Y|X=0,z)\prob(z|X=0) \right].
\end{eqnarray*}
When $\prob(z|X)\neq\prob(z)$, these quantities do not coincide.
However, any method that can estimate $\prob(Y|X,Z)$ for all levels of $Z$ and $X$ will 
arrive at the correct ATT estimate.  We used logistic regression in our implementation
of the Direct Adjustment method, which can estimate $\prob(Y|X,Z)$ when $X$ and $Z$
have no interactions. Note that the causal graph admits interaction between $X$ and $Z$,
thus model misspecification can cause biases in the estimate.

\experiment{IV. Confounding with Indirect Effect.}
In addition to the Confounding experiment, we also have an indirect causal effect from $U$.
We now have two diseases, $Z$ and $U$, each of which can be treated with $X$.
20\% of the population has $Z$ and independently, 20\% has $U$.
25\% of the patients who have $Z$ and have no treatment ($X=0$) have $Y$,
while only 10\% of the treated ($X=1$) patients have it, regardless of whether
the patient has $U$. (If the probability of $Y$ was affected by $U$, it would be another
confounder, rather than have an indirect effect.) \\
$X$ has a beneficial ATT of -.15 in patients with $Z==1$ (and $X==1$) and has no effect
in patients with $Z=0$ (who get $X$ because of $U$). Thus the true ATT=-.0833. \\
In this experiment, the counterfactual model was the best-performing model.
The counterfactual model estimates the ATT through the definition
\begin{displaymath}
\att = \E\left[ \prob(Y_1|X=1) - \prob(Y_0|X=1)\right],
\end{displaymath}
where $Y_0$ is the potential outcome the patient would have without treatment $X=0$
and $\prob(Y_0|X=1)$ is the counterfactual probability of $Y$ (the probability of $Y$
had they not received $X$) in the population who actually got $X=1$.  
Note that the potential outcome $Y_1|X=1$ in
the patients who actually got $X=1$ is the observed outcome $Y|X=1$.
With $u$ and $z$ denoting the levels of $U$ and $Z$, respectively and
$\prob(u)$ being a shorthand for $\prob(U=u)$,
\begin{eqnarray*}
ATT &=& \E\left[ \prob(Y|X=1) - \prob(Y_0|X=1)\right] \\
&=& \sum_u \sum_z \prob(u,z) \left[ \prob(Y|X=1,u,z)-\prob(Y_0|X=1,u,z) \right] \\
&=& \sum_z \prob(z) \sum \left[ \prob(Y|X=1,z)-\prob(Y_0|X=1,z) \right] \\
&=& \sum_z \prob(z) \sum \left[ \prob(Y|X=1,z)-\prob(Y|X=0,z) \right], 
\end{eqnarray*}
which coincides with the data generation mechanism, hence the estimate is correct. \\

In the derivation, step 2 holds because $U$ and $Z$ are independent given $X$
and step 3 uses the fact that the counterfactual model estimates $P_0(Y|X=1,z,u)$ 
from the untreated patients, thus
\begin{displaymath}
\prob(Y_0|X=1,z,u)=\prob(Y|X=0,z,u)=\prob(Y|X=0,z).
\end{displaymath}
\experiment{V. Confounding with Direct and Indirect Effects.}
In this experiment, we have three diseases: our index disease $Z$, which is a confounder;
$U$ having an indirect effect on $Y$ via $X$; and $V$ having a direct effect on $Y$.
20\% of the population has each of $Z$, $V$ and $U$ independently.
95\% of patients with $Z$ or $U$ get the intervention $X$.
25\% of the untreated patients with $Z$ get $Y$, while only 10\% of the treated patients do,
regardless of whether they have $U$.
Patients with $V$ face a 5\% in their chance of experiencing outcome $Y$. \\
$X$ has a beneficial ATT of -.15 in patients with $Z=1$ and have no effect in patients with 
$Z=0$ (who get $X$ because of $U$). Whether a patient has $V$ does not influence the
effect of $X$.  The true ATT is thus -.0833. \\
None of the methods estimated the effect correctly, but Propensity Score Matching
came closest.
Analytic derivation of why it performed well is outside the scope of this paper,
but in essence, its success is driven by its ability to maximally exploit the independence
relationships encoded in the causal graph.  It can ignore $V$ when it constructs
the propensity score model, because $X$ and $V$ are  independent
(when $Y$ not given); and it can ignore $U$ and $V$ when it computes the 
ATT in the propensity matched population.  
On the other hand, the causal graph admits interaction among $U$, $Z$ and $X$,
thus a logistic regression model as the propensity score model can be subject
to model misspecification. \\
The Stratified Non-Parametric method, which is essentially just a direct implementation
of the definition of ATT, underestimated the ATT by almost 25\%.  The reason lies
in the excessive stratification across all combinations of the levels of $U$, $V$, and $Z$. 
Even with just three variables, most strata did not have sufficiently many patients 
(either treated or untreated) to estimate $\prob(Y|X,u,v,z)$.  In the discussion, 
we will describe remedies to overcome this problem.
\begin{table}[!ht]
%\small
\begin{tabular}{lrrrrrr}
\hline
      & Conf & CC    &DA       &    CM  &  PSM & SN \\
\hline
I.   & +.110 & -.150 & -.150 &   -.150 &      NA &   -.150 \\
II.  & +.099 & -.150 & -.150 &   -.150 &   -.151 &  -.149\\
III. & +.099 & +.047 & -.136 &  -.136 &   -.136 &  -.136 \\
IV. & +.077 & +.024 & -.019 &  -.083 &   -.068 &  -.064 \\
V.  & +.072 & +.038 & -.037 &  -.105 &   -.074 &  -.067 \\
\hline
\end{tabular}
\caption{The ATT estimates by the 6 methods in the five experiments. The experimental conditions,
the full names of the methods and the true ATT value are specified in the text.}\label{tbl:synthetic}
\end{table}

\section{Discussion And Conclusion}
We proposed the causal rule mining framework, which
transitions pattern mining from finding patterns that are associated with an outcome towards patterns
that cause changes in the outcome. Finding causal relationships instead of associations is absolutely
critical in health care, but also has appeal beyond health care.

The numerous biases that arise in establishing causation make quantifying causal effects difficult.
We use the Neyman-Rubin causal model to define causation and use the potential outcome framework
to estimate the causal effects. We correct for three kinds of potential biases: those stemming from
direct causal effect, indirect causal effect and confounding. We compared five different methods
for estimating the causal effect, evaluated them on real and synthetic data and found that three
of these methods gave very similar results.

We have demonstrated on real clinical data that our proposed method can effectively enumerate
causal patterns in a large combinatorial search space due to the two new pruning methods
we developed for this work.  We also demonstrated that the patterns discovered from the
data were very rich and we managed to illustrate how the effect of statin is different in various
subpopulations. The results we found are consistent with the literature but go beyond what is
already known about statin's effect on the risk of diabetes.

The discussions and experimental results provided in this paper provide some general guidance on when to use the different methods we described. We recommend counterfactual confidence if no confounding is suspected as
counterfactual confidence is computationally efficient and can arrive at the correct solution even when
direct effects and indirect effects are present. In the presence of confounding, propensity score matching gave the most accurate results, but due to the need to create a matched population, it has built-in randomness, increasing its variance. Moreover, the counterfactual model as well as the propensity score model are susceptible to model misspecification. If unknown interactions among variables are suspected, we recommend the stratified non-parametric method. With this technique,
model misspecification is virtually impossible, however, its sample size requirement is high.
The stratified model is suboptimal if we need to stratify across many variables.
Stratifying across many variables can  fragment the population into many strata too small to
afford us with the ability to estimate the effects correctly.  If the estimates use some strata but
not others, they may be biased.
\section*{Acknowledgments}
This study is supported by National Science Foundation (NSF) grant: IIS-1344135. Contents of this document are the sole responsibility of the authors and do not necessarily represent
official views of the NSF. 
\bibliography{sigproc.bib}
\bibliographystyle{unsrt}
\end{document}